\documentclass[lettersize,journal]{IEEEtran}
\usepackage{amsmath,amsfonts}
\usepackage{algorithmic}
\usepackage{array}
\usepackage[caption=false,font=normalsize,labelfont=sf,textfont=sf]{subfig}
\usepackage{textcomp}
\usepackage{stfloats}
\usepackage{url}
\usepackage{algorithm}
\usepackage{hyperref}
\usepackage{verbatim}
\usepackage{graphicx}
\usepackage{pdfpages}
\usepackage{booktabs}
\usepackage{amssymb}
\usepackage{array}
\usepackage{amsmath}
\usepackage{listings}
\usepackage{utfsym}
\usepackage{xcolor}
\usepackage{float}
\usepackage{colortbl}
\usepackage{multirow}
\usepackage{authblk}
\usepackage{pifont}
\usepackage[T1]{fontenc}
\usepackage{amsmath,bm}
\usepackage{makecell} 
\usepackage{mathrsfs}
\usepackage{booktabs} 
\usepackage{multirow} 
\usepackage{array}    
\usepackage{threeparttable}
\usepackage[numbers,sort&compress]{natbib}
\def\BibTeX{{\rm B\kern-.05em{\sc i\kern-.025em b}\kern-.08em
    T\kern-.1667em\lower.7ex\hbox{E}\kern-.125emX}}
\usepackage{balance}
\begin{document}
\title{\LARGE Few-Shot Identity Adaptation for 3D Talking Heads via \\ Global Gaussian Field}
\author{Hong Nie, Fuyuan Cao, Lu Chen, Fengxin Chen, Yuefeng Zou, Jun Yu$^*$
\thanks{
$^*$ Corresponding author: Jun Yu (harryjun@ustc.edu.cn).
}}

\markboth{}%
{How to Use the IEEEtran \LaTeX \ Templates}

\maketitle

\begin{abstract} 
Reconstruction and rendering-based talking head synthesis methods achieve high-quality results with strong identity preservation but are limited by their dependence on identity-specific models. Each new identity requires training from scratch, incurring high computational costs and reduced scalability compared to generative model-based approaches. To overcome this limitation, we propose FIAG, a novel 3D speaking head synthesis framework that enables efficient identity-specific adaptation using only a few training footage. FIAG incorporates Global Gaussian Field, which supports the representation of multiple identities within a shared field, and Universal Motion Field, which captures the common motion dynamics across diverse identities. Benefiting from the shared facial structure information encoded in the Global Gaussian Field and the general motion priors learned in the motion field, our framework enables rapid adaptation from canonical identity representations to specific ones with minimal data. Extensive comparative and ablation experiments demonstrate that our method outperforms existing state-of-the-art approaches, validating both the effectiveness and generalizability of the proposed framework. Code is available at: \textit{https://github.com/gme-hong/FIAG}.
\end{abstract}



\definecolor{mypink}{HTML}{EEA9AD}
\definecolor{myyellow}{HTML}{EFF1AF}
\definecolor{mygreen}{HTML}{BADBBD}

\section{Introduction}
\IEEEPARstart{T}{he} technology of talking head synthesis has gained widespread applications across diverse domains, including video production \cite{zakharov2019few, yi2020audio, chen2020comprises, zhou2020makelttalk, zhang2021text2video, gao2023high, shen2023difftalk}, virtual reality \cite{christoff2023application, gunawardhana2024toward}, language learning \cite{anderson2008language, gu2022systematic}, human-computer interaction \cite{zhen2023human, delbosc2023towards} and so on. With ongoing technological advancements, methodologies for talking head synthesis have transitioned from dependence on intricate motion capture systems \cite{hsieh2015unconstrained} to contemporary approaches grounded in deep generative models \cite{song2018talking, jamaludin2019you, chen2019hierarchical, kr2019towards, prajwal2020lip, wang2021audio2head, zhang2021facial, ji2021audio, singer2022make, yu2023talking, stypulkowski2024diffused}. These approaches generate realistic talking head videos from only a single reference image and driving audio. However, generative models often exhibit suboptimal performance regarding image clarity, identity preservation, and the ability to accommodate extreme head movements. The advent of 3D reconstruction techniques like Neural Radiance Fields (NeRF) \cite{mildenhall2021nerf} and 3D Gaussian Splatting (3D GS) \cite{kerbl20233d}, known for fast reconstruction and high-quality rendering, offers promising solutions to challenges faced by generative models in talking head synthesis. Numerous works \cite{gafni2021dynamic, guo2021ad, liu2022semantic, hong2022headnerf, yao2022dfa, qian2024gaussianavatars, dhamo2024headgas, xu2024gaussian} have advanced dynamic reconstruction by disentangling representation and motion fields, laying a foundation for applying NeRF and 3D GS to talking head synthesis.

\begin{figure}[t]
\centering
\includegraphics[width=\linewidth]{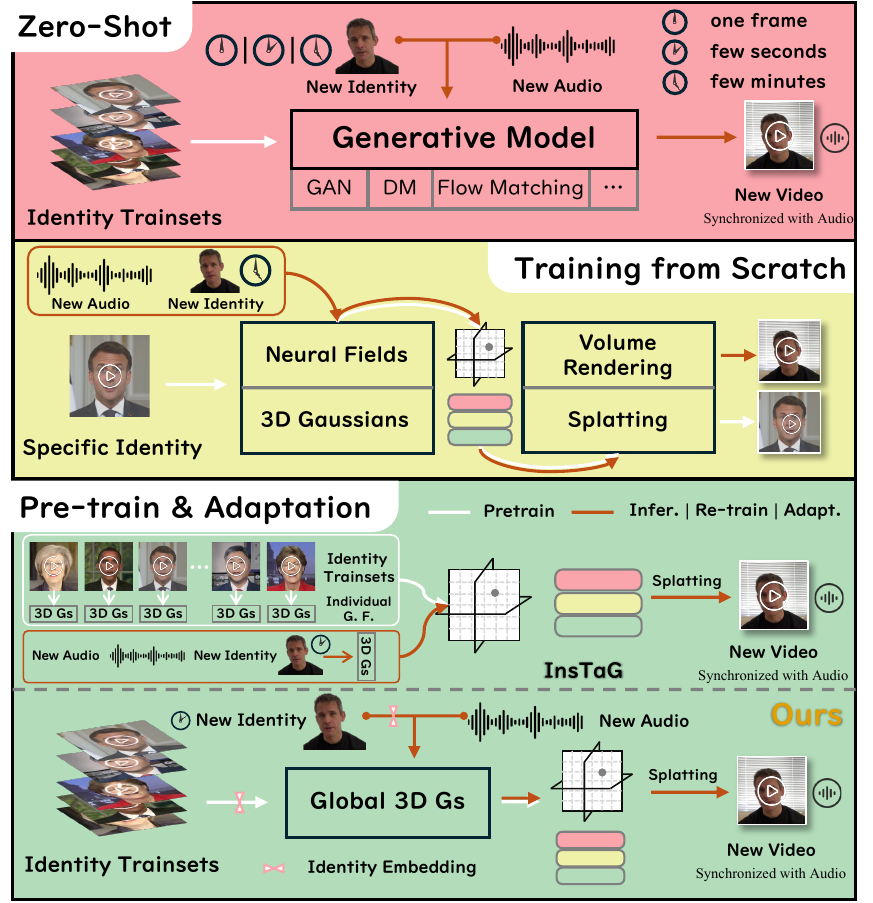}
\caption{\textbf{Comparison of various frameworks for talking heads generation.} (1) the zero-shot generative approach (top), which employs generative models with firsthand inference of new identities; (2) the RAR-based method (middle), utilizing identity-specific models for precise head reconstruction; and (3) the PAA paradigm (bottom), adopting multi-identity pretraining followed by targeted finetuning for novel identities.}
\label{fig1}
\end{figure}

Recent studies \cite{shen2022learning, tang2022real, ye2023geneface, ye2023geneface++, shen2023sd, li2024talkinggaussian, cho2024gaussiantalker, liang2024cstalk} have demonstrated that reconstruction and rendering (RAR)-based talking head synthesis mitigates some limitations of generative models. However, these methods rely on identity-specific representation fields, restricting NeRF and 3D GS models to individual identities. Hence, they require training from scratch for new identities, resulting in inefficient generalization compared to generative approaches. This inefficiency is especially severe for NeRF models, demanding days of training, whereas 3D GS reduces to hours.

To mitigate the inefficiency of training from scratch for each identity, several works \cite{ye2024mimictalk, li2025instag} adopt a pretraining and adaptation (PAA) paradigm, enabling finetuning on new identities by leveraging pretrained knowledge. This involves learning a universal motion field during pretraining, which is then adapted to identity-specific motions. While this improves adaptation efficiency, limitations persist: the representation field remains identity-exclusive and must be rebuilt for each identity, requiring separate fields in both pretraining and adaptation. Additionally, these methods neglect shared topological structures of facial features across identities despite recognizing common motion patterns.

To address these challenges, we propose \textbf{FIAG}, which achieves \textbf{F}ew-shot \textbf{I}dentity \textbf{A}daptation for 3D talking heads via \textbf{G}lobal Gaussian Field, extending the PAA paradigm \cite{ye2024mimictalk, li2025instag} by finetuning both the representation and motion fields, as shown in Fig. \ref{fig1}. Our Global Gaussian Field enables multi-identity pretraining and identity-specific adaptation within a unified representation, substantially reducing computational and storage costs. Leveraging shared head topology across identities further simplifies adaptation for new identities.

A pivotal challenge to achieve the aforementioned scheme is that Gaussian fields are typically identity-specific, causing conflicts if shared across multiple identities. To overcome this, our Global Gaussian Field is decomposed into a shared Gaussian field and an identity embedding module. The module extracts identity-specific features that, combined with the Gaussian ellipsoids, are input to MLPs to activate relevant ellipsoids and infer identity-specific offsets. Extensive comparative and ablation studies validate the effectiveness of FIAG.

The main contributions of this work are summarized as follows:
\begin{enumerate}
  \item To the best of our knowledge, we are the first to employ a Global Gaussian Field for PAA across multiple identities within a unified representation, effectively overcoming Gaussian field exclusivity and resolving identity conflicts arising from shared usage.

  \item The success of the Global Gaussian Field in the talking head synthesis task provides a solution for other similar tasks. Namely, the representation fields of "different identities" with common traits could be decoupled into a global representation \& "identity embedding" pattern.

  \item Extensive comparative and ablation studies demonstrate that our method outperforms state-of-the-art approaches and validates its effectiveness.
  
\end{enumerate}

\section{Related work}
\subsection{Identity-Agnostic Talking Head Synthesis}
The identity-agnostic paradigm in talking head synthesis focuses on motion learning while masking speaker-specific traits, typically via large-scale training on diverse identity-rich videos to enable generalized motion reasoning for unseen identities. This approach mainly uses generative architectures such as GANs \cite{10.1145/3343031.3351066, 8953690, prajwal2020lip}, Diffusion Models \cite{shen2023difftalk, stypulkowski2024diffused, li2024latentsync}, and Flow Matching \cite{ki2024float}. For example, Wav2Lip \cite{prajwal2020lip} employs a GAN with a lip-sync discriminator for audiovisual alignment; Stypułkowski et al. \cite{stypulkowski2024diffused} use an autoregressive diffusion model for photorealistic video synthesis from a single identity image and audio; FLOAT \cite{ki2024float} applies flow-matching for temporally coherent audio-driven portraits. While these methods enable zero-shot inference on unseen identities without model retraining, a significant practical advantage, their identity-agnostic formulation intrinsically limits speaker consistency preservation. Moreover, despite substantial progress in generative fidelity, persistent challenges remain in output quality, particularly regarding subtle facial dynamics and the mitigation of perceptible synthesis artifacts.

\subsection{Identity-Specific Talking Head Synthesis}

Identity-specific talking head synthesis constructs subject-exclusive models trained on individual videos, typically using 3D reconstruction techniques such as NeRF \cite{guo2021ad, yao2022dfa, tang2022real, ye2023geneface, li2023efficient, peng2024synctalk, ye2024mimictalk} and 3D GS \cite{li2024talkinggaussian, cho2024gaussiantalker, aneja2024gaussianspeech, deng2025degstalk, cha2025emotalkinggaussian, li2025instag}. For instance, Guo et al. \cite{guo2021ad} map audio features directly to NeRF via a conditional implicit function, modeling head and upper body with dual volume renderers, while Li et al. \cite{li2024talkinggaussian} deform persistent Gaussian primitives in 3D GS with a dual-branch design to decouple face-mouth motion for precise lip-sync and structural fidelity. Although these methods ensure strong identity consistency, their per-identity training is inefficient. Recent approaches \cite{ye2024mimictalk, li2025instag} address this by pretraining a universal motion field on multi-identity data and finetuning it for new identities, reducing adaptation to seconds of footage. Building on PAA, our method transforms single-identity Gaussian fields into a multi-identity framework via shared Gaussian parameters and identity embeddings, enhancing parameter reuse and computational efficiency.

\section{Methodology}

\begin{figure*}[t]
\centering
\includegraphics[width=\textwidth]{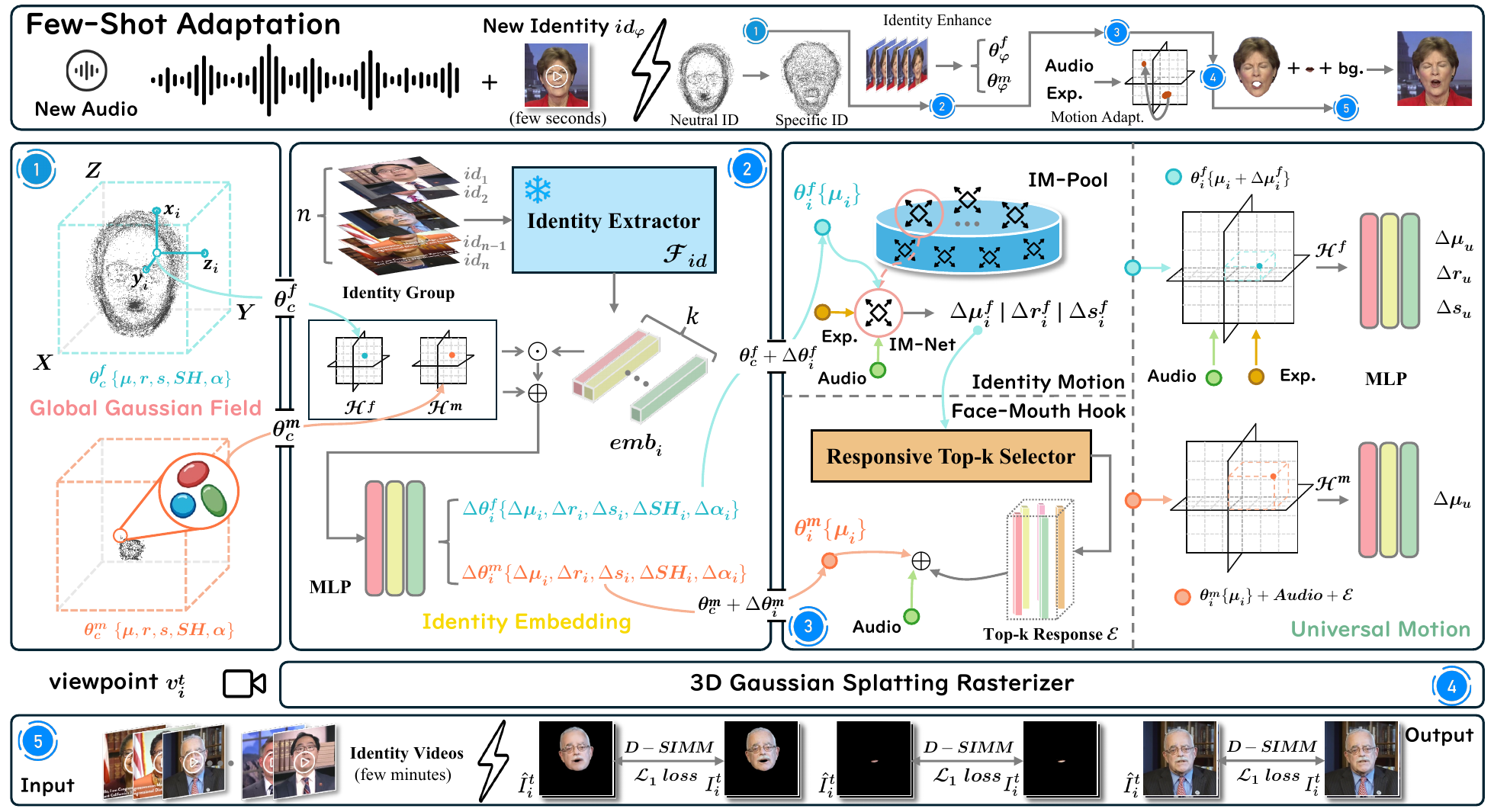}
\caption{\textbf{Overview of the proposed FIAG.} In pretraining, FIAG constructs a Global Gaussian Field to encode canonical head features across multiple identities, with an identity embedding module resolving inter-identity conflicts, while learning generalized motion patterns stored in a Universal Motion Field. During finetuning on sparse footage of new idenities, the framework transforms neutral-to-specific representations via the Gaussian field to capture identity-discriminative details, and adapts the pre-trained motion field for personalized dynamics, enabling high-fidelity talking head synthesis with precise identity preservation.}
\label{fig2}
\end{figure*}

\begin{figure}[t]
\centering
\includegraphics[width=\linewidth]{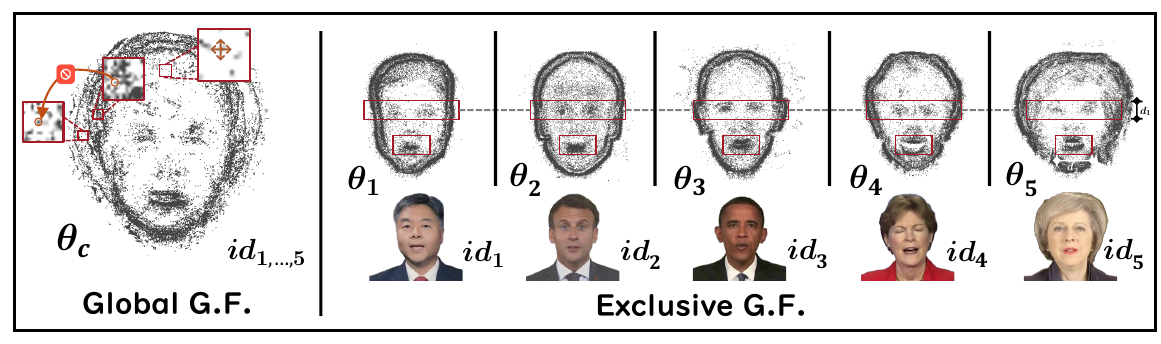}
\caption{\textbf{The distinction between an Exclusive Gaussian Field and a Global Gaussian Field.} The global Gaussian field models the majority of common head features (e.g., geometric contours and topological structures) while minimizing the emphasis on features specific to a particular identity.}  
\label{fig3}
\end{figure}

As illustrated in Fig. \ref{fig2}, our proposed FIAG framework builds upon 3D Gaussian Splatting \cite{kerbl20233d}. The model architecture comprises four key components: Global Gaussian Field for a canonical static spatial representation, Identity Embedding module to acquire subject-specific spatial representation by capturing identity feature embedding, Motion Field modeling dynamic transformations, and differentiable Rasterizer for rendering.
\subsection{Global Gaussian Field}

In the context of the talking heads generation, Gaussian Field serves as a fundamental representation framework, where a set of parametric Gaussian ellipsoids collectively encodes the static identity characteristics of a target subject. Each ellipsoid \( g_i = \{\mu_i, r_i, s_i, SH_i, \alpha_i\} \) encapsulates spatially geometrical attributes by Eq. (\ref{exp1}) and photometric properties for rendering through spherical harmonic coefficients and opacity value.
\begin{equation}
\label{exp1}
g(x)= exp(-\frac{1}{2}(x-\mu)^T\Sigma^{-1}(x-\mu)),
\end{equation}
where $\mu \in \mathbb{R}^3$ represents the center of the ellipsoid, and the covariance matrix $\Sigma$ is calculated through the quaternion $r \in \mathbb{R}^4$ and the scaling factor $s \in \mathbb{R}^3$. The composite field \(\theta = \{g_i\}_{i=1}^n\) forms a completely differentiable representation that bridges discrete morphological descriptors with continuous deformation spaces, enabling both explicit geometric manipulation and implicit neural rendering paradigms. 

When modeling for a single specific identity, a single Gaussian field suffices. However, for multi-identity modeling, it becomes necessary to construct distinct Gaussian fields $\theta_i \{\mu_i, r_i, s_i, SH_i, \alpha_i\}$ \cite{li2025instag} for each identity $i$ to store their respective static attributes, as shown in Fig. \ref{fig2} (1). To maximize the utilization efficiency of Gaussian ellipsoids, we propose a global Gaussian field \(\theta_c \{\mu, r, s, SH, \alpha\}\) across all identities. The core principle involves reusing existing ellipsoids through a parametric offset \(\Delta \theta\) that transforms canonical parameters \(\theta_c\) to identity-specific instances \(\theta_i\). For cases where excessive positional deviations may induce motion gradient explosion, new ellipsoids are dynamically instantiated to ensure numerical stability. Based on this principle, the global Gaussian field intentionally attenuates subtle individual-specific features (e.g., fine facial wrinkles) that lack universality across identities. This attenuation effect becomes more pronounced as the number of training identities increases. Consequently, the objective of the global Gaussian field during pretraining is to model macro-level features that sufficiently characterize identity representation, rather than capturing microscopic particulars, while the refinement of detailed features is delegated to the adaptation stage. Besides, finetuning novel identities directly within the pretrained global Gaussian field, which inherently encapsulates shared facial attributes (e.g., facial feature topology, presence of eyewear) across multiple identities, exhibits significantly higher efficiency than de novo modeling of new identities.

Consistent with previous methodologies \cite{cho2024gaussiantalker, deng2025degstalk, li2024talkinggaussian}, we adopt a segmentation strategy for facial and oral regions. This approach serves a dual purpose: 1) enhancing the reconstruction of intraoral details, and 2) mitigating the inconsistency in motion granularity between facial expressions and internal oral movements \cite{li2024talkinggaussian}.

\subsection{Identity Embedding}
Employing a single Global Gaussian Field to model multiple identities would inevitably induce severe identity conflicts, as extensively analyzed by Li \textit{et al.} \cite{li2025instag}. Two critical issues emerge:

\begin{enumerate}
    \item Gaussian Allocation: The determination of which Gaussians within the Global Gaussian Field should be allocated to a specific identity $i$;
    \item Parameter Displacement: Quantifying the exact offset value of parameters $\Delta \theta$ required to composite the complete feature set of a target identity $i$.
\end{enumerate}
These challenges can be formally expressed as:
\begin{equation}
\label{exp2}
\theta_i = \theta_s + \Delta \theta, \quad \theta_s \subseteq \theta_c.
\end{equation}

Although the aforementioned problems appear to be classification and regression tasks respectively, they can be implicitly reformulated as generation problems for unified resolution. As illustrated in Fig. \ref{fig2} (1, 2), the proposed method leverages a tri-plane hash encoder $\mathcal{H}$ to transform all Gaussian ellipsoids within the Global Gaussian Field from Euclidean space representation into a high-dimensional latent space. Subsequently, identity-specific embeddings $emb_i$ are extracted for each subject via a pretrained identity feature extractor $\mathcal{F}_{id}$. These latent geometric features and identity embeddings are then fused and decoded back into the original space through MLPs, yielding an identity-conditioned subset of activated Gaussian primitives along with corresponding parameter offsets. 

The process can be formulated as:
\begin{equation}
\label{exp3}
\Delta \theta = MLP(\mathcal{H}(\theta_c\{\mu\}) \oplus (\mathcal{H}(\theta_c\{\mu\}) \odot \mathcal{F}_{id}(id_i))).
\end{equation}

\begin{figure}[t]
\centering
\includegraphics[width=\linewidth]{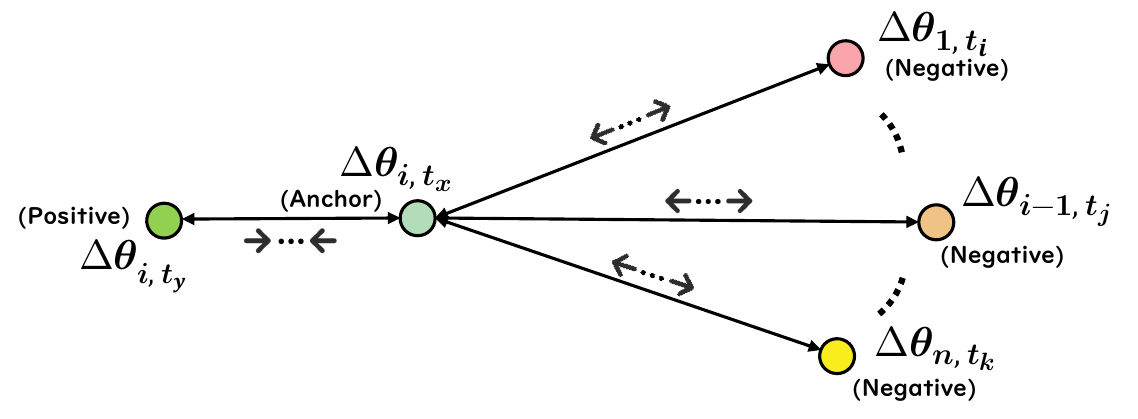}
\caption{\textbf{Triplet loss for identity isolation.} The triplet loss operates by constructing: (1) an anchor (identity $i$'s current offset), (2) a positive sample (same identity's temporal offset), and (3) a negative sample (cross-identity offset). This formulation guides the network to resolve Gaussian Allocation by maximizing inter-identity margin while minimizing intra-identity variance, thereby achieving precise identity isolation in the feature space.}
\label{fig4}
\end{figure}

The identity feature embedding \( emb_i \) already provides identity-specific cues to the MLP network. To further clarify the optimization direction, we introduce a triplet loss \( \mathcal{L}_t \) \cite{schroff2015facenet}. This loss optimizes the distances among triplets (Anchor, Positive, Negative) to enhance identity disentanglement, as illustrated in Fig. \ref{fig4}.
\begin{equation}
\label{exp4}
\mathcal{L}_t = \sum_i^n\Big[||\Delta \theta_{i}^a - \Delta \theta_{i}^p ||^2 - ||\Delta \theta_{i}^a -\Delta \theta_{\phi}^n ||^2 + \tau \Big]_+,
\end{equation}
where $\Delta \theta_{i}^a$ represents the \textbf{a}nchor of the parameter offset for identity $i$, $\phi = [n]\textbackslash \{i\}:= \{1, \ldots, n\}\textbackslash \{i\}$, and $\tau$ denotes the enforced margin between positive and negative sample pairs.

By integrating both explicit (Global Gaussian Field) and implicit (Identity Embedding) representation paradigms, the Identity Embedding stage yields a static Gaussian field that is specific to each individual identity.

\subsection{Motion Field}
Gaussian fields are fundamentally limited to static structural representations, while talking heads generation requires modeling temporal dynamics. The motion field is specifically designed to capture this dynamic process. Following the established paradigm in prior works \cite{guo2021ad, yao2022dfa, tang2022real, shen2022learning, li2023efficient, ye2023geneface, peng2024synctalk, ye2024mimictalk, cho2024gaussiantalker, li2024talkinggaussian, deng2025degstalk, li2025instag, aneja2024gaussianspeech, cha2025emotalkinggaussian}, for identity instance $i$, the goal is to synthesize facial expression motion \(\delta_i^f \{\Delta \mu_i^f, \Delta r_i^f, \Delta s_i^f\}\) and lip motion \(\delta_i^m \{\Delta \mu_i^m\}\) conditioned on driving signals \(\mathcal{C}\) (e.g., audio, expression):
\begin{equation}
\label{exp4}
\delta_i = \mathcal{M}(\theta_i\{\mu_i\}, C) = MLP(\mathcal{H}(\theta_i\{\mu_i\} \oplus C)),
\end{equation}
where $\mathcal{M}$ represents the motion field, and $\theta_i\{\mu_i\}$ denotes the center of Gaussian ellipsoids belonging to a specific identity $i$.
To minimize conflicts between different motion attributes of multiple identities while avoiding excessive additional motion fields and preserving the coupling between facial and intra-oral movements, we partition the motion field into three key components: the Identity Motion Field, Face-Mouth Hook Module, and Universal Motion Field.

\subsubsection{Identity Motion}
The Identity Motion Field is designed to capture personalized facial dynamics through the initialization of an Identity-specific Motion Pool (IM-Pool), which consists of lightweight Identity Motion Networks (IM-Net), each uniquely associated with a particular identity. The networks process three key input parameters: face branch parameters, corresponding facial Action Units \cite{prince2015facial} (AUs), and accompanying audio signals, subsequently generating predicted motion variations. When producing personalized facial movements for identity i, we retrieve the respective IM-Net from IM-Pool to compute the requisite motion predictions $\delta_i^f$, as illustrated in the upper-left quadrant of Fig. \ref{fig2} (3). The Negative Contrast Loss \cite{li2025instag} is employed to guide the learning process of this procedure.
\begin{equation}
\label{exp6}
\mathcal{L}_c = \mathbb{I}_{trunc} (\theta_i^f\{\mu_i + \Delta \mu_i^f\} \cdot \theta_{\phi}^f\{\mu_{\phi} + \Delta \mu_\phi^f\})
\end{equation}

\subsubsection{Face-Mouth Hook}
While constructing a dedicated motion field solely for the mouth region would be computationally prohibitive, the mouth component remains intrinsically linked to facial movements. To resolve this challenge, we adopt the Face-Mouth Hook (Li \textit{et al.} \cite{li2025instag}), where personalized facial motion outputs are processed by a Top-k selector to extract critical motion features (hook $\mathcal{E}$) from the face branch, which are then integrated into the universal motion pipeline of the mouth branch, as shown in the lower-left quadrant of Fig. \ref{fig2} (3).

\subsubsection{Universal Motion}
The essence of audio-driven facial animation lies in establishing a mapping between audio features and lip movements \cite{peng2024synctalk}. However, in-the-wild monologue videos of a single identity are often limited in duration, making it challenging for single-identity training to effectively cover the majority of potential mappings. Consequently, multi-identity training enables the learning of broader mapping relationships, thereby facilitating more accurate lip synchronization during inference.  The universal motion field $\mathcal{U}$ aims to establish shared mapping relationships across multiple identities. As shown in Fig. \ref{fig2} (3), for the face branch, it takes the identity motion output $\theta_i^f\{\mu_i + \Delta \mu_i^f\}$ of identity $i$, along with audio and expression actions as inputs, and generates a universal facial motion offset $\delta_c^f\{\Delta \mu_u, \Delta r_u, \Delta s_u\}$ through a tri-plane encoder followed by MLPs decoder. 
\begin{equation}
\label{exp7}
\delta_i^f = \mathcal{U}(IM\text{-}Net(\theta_i^f\{\mu_i\}, C), C), C = Audio + Exp.
\end{equation}
Meanwhile, the mouth branch utilizes the Gaussian parameters $\theta_i^m\{\mu_i\}$ of identity $i$, audio features, and hook $\mathcal{E}$ as inputs to predict a universal lip motion offset $\delta_c^m\{\Delta \mu_u\}$ .
\begin{equation}
\label{exp8}
\delta_i^m = \mathcal{U}(\theta_i^m\{\mu_i\}, Audio, \mathcal{E}).
\end{equation}

\subsection{Rasterizer}
As shown in Fig. \ref{fig2} (4, 5), the rasterizer $\mathcal{R}$ utilizes the differential splatting \cite{yifan2019differentiable} to render the 3D Gaussian ellipsoids into a 2D image $\hat{I}_i^t$ from a specific viewpoint $v_i^t$ for a given identity $i$:
\begin{equation}
\label{exp9}
\mathcal{R}: (\theta_i + \delta_i, v_i^t) \rightarrow \hat{I}_i^t,
\end{equation}
where the face $(\theta_i^f + \delta_i^f)$ and mouth $(\theta_i^m + \delta_i^m)$ branches are respectively rendered to form the final image of the talking head. For the rendered images $\hat{I}$, we employ both $\mathcal{L}_1$ loss and D-SSIM loss to measure their discrepancy with the ground truth images \( I \):
\begin{equation}
\label{exp9}
\mathcal{L} = \lambda_1 \mathcal{L}_1(\hat{I}, I) + \lambda_2 D\text{-}SSIM(\hat{I}, I).
\end{equation}

\subsection{Few-Shot Adaptation}
The adaptation to a new identity \(\varphi\) requires only seconds of video footage for finetuning, significantly less than the minutes-long sequences needed during pretraining. Crucially, this identity adaptation process leverages most existing modules from the pretraining framework (e.g., Global Gaussian Field and Universal Motion Fields), requiring only the additional initialization of an IM-Net. As illustrated in Fig. \ref{fig2} (top), the adaptation process progressively transforms the neutral-oriented global Gaussian representation into an identity-specific representation, where facial details are gradually reconstructed. The identity embedding module further enhances identity consistency, while the newly initialized IM-Net collaborates with the pre-trained universal motion field to achieve identity-adaptive motion modulation. This pipeline ultimately generates high-quality talking head images with both strong identity preservation and rich detail reproduction.

\begin{table*}[t]
\centering
\caption{Comparison of various methods under different frameworks with self-reconstruction configuration across three critical dimensions: image quality, lip-sync accuracy, and generation efficiency.}
\label{tab1}
\renewcommand{\arraystretch}{1.3}
\begin{tabular}{lcccccccccc}
\toprule
\multirow{2}{*}{\centering\vspace*{-6pt} Methods} & \multirow{2}{*}{\centering\vspace*{-6pt}\centering Config.} & \multicolumn{3}{c}{Image Quality} & \multicolumn{3}{c}{Lip-sync Accuracy} & \multicolumn{2}{c}{Efficiency} & \multirow{2}{*}{\centering\vspace*{-6pt}  Real-time} \\
\cmidrule(lr){3-5} \cmidrule(lr){6-8} \cmidrule(lr){9-10}
 & & PSNR $\uparrow$ & LPIPS $\downarrow$ & SSIM $\uparrow$ & LMD $\downarrow$ & AUE-(U/L) $\downarrow$ & Sync-C $\uparrow$ & Training $\downarrow$ & FPS $\uparrow$ \\
\midrule
Ground Truth & - & - & - & - & - & - & 2.023 & - & - & -\\
\hline
Wav2Lip \cite{prajwal2020lip} & \multirow{3}{*}{\centering Zero-shot} & 33.3163 & 0.0612 & 0.9480 & 2.6447 & - / 0.512 & \underline{2.305} & - & 22.6 & \color{red}\ding{56} \\
LatentSync \cite{li2024latentsync} &  & 33.3389 & 0.0330 & 0.9535 & 3.1159 & - / 1.079 & 1.856 & - & 0.6 & \color{red}\ding{56} \\
FLOAT \cite{ki2024float} &  & 17.4075 & 0.3373 & 0.5774 & 3.8599 & 1.530 / 1.486 & \textbf{6.067} & - & 33.3 & \color{green}\ding{51} \\
\hline
EAT \cite{gan2023efficient} & \multirow{2}{*}{\centering One-shot} & 18.9929 & 0.3775 & 0.6270 & 5.0353 & 2.824 / 9.442 & 0.788 & - & 10.2 & \color{red}\ding{56} \\
Real3DPortrait \cite{yereal3d} & & 19.1395 & 0.3554 & 0.6412 & 5.1698 &  1.276 / 1.572 & 0.281 & - & 7.6 & \color{red}\ding{56} \\
\hline
DFRF \cite{shen2022learning} & \multirow{5}{*}{\centering \makecell{Training \\ from \\ Scratch}} & 18.7434 & 0.2439 & 0.6373 & 4.8127 & 1.369 / 2.308 & 0.801 & 9 hours & 0.03 & \color{red}\ding{56}\\
RAD-NeRF \cite{tang2022real} & & 26.4932 & 0.0975 & 0.8415 & 2.3206 & 0.214 / 0.292 & 1.340 & 5 hours & 27.2 & \color{green}\ding{51} \\
ER-NeRF \cite{li2023efficient} & & 25.7446 & 0.0778 & 0.8438 & 2.2220 & 0.250 / 0.221 & 1.764 & 2 hours & 28.8 & \color{green}\ding{51}\\
TalkingGaussian \cite{li2024talkinggaussian} & & \underline{36.2547} & \textbf{0.0207} & \underline{0.9677} & \underline{2.1397} & 0.314 / \textbf{0.087} & 1.710 & 36 mins & \textbf{112.4} & \color{green}\ding{51} \\
GaussianTalker \cite{cho2024gaussiantalker} & & 34.7823 & 0.0427 & 0.9566 & 2.2982 & \underline{0.146} / 0.447 & 1.718 & 75 mins & 61.8 & \color{green}\ding{51}\\
\hline
MimicTalk \cite{ye2024mimictalk} & \multirow{3}{*}{\centering \makecell{Pre-train \& \\ Adaptation}}  & 24.1083 & 0.1088 & 0.7686 & 2.5822 & 0.962 / 0.867 & 1.703 & 24 mins & 8.2 & \color{red}\ding{56}\\
InsTaG \cite{li2025instag} & & 35.8987 & 0.0375 & 0.9610 & 2.1945 & 0.156 / 0.286 & 1.741 & \underline{19 mins} & \underline{80.3} & \color{green}\ding{51} \\
\textbf{FIAG (Ours)} & & \textbf{37.6899} & \underline{0.0296} & \textbf{0.9716} & \textbf{2.0954} & \textbf{0.111} / \underline{0.174} & 1.909 & \textbf{16 mins} & 65.5 & \color{green}\ding{51} \\
\bottomrule
\end{tabular}
\end{table*}

\section{Experiments}

\subsection{Experimental Setup}
All experiments are conducted on NVIDIA RTX 3090 GPUs, with the total pretraining time approximately 2 hours. Action Unit Error \cite{prince2015facial} (AUE) comprises upper-face action unit error (AUE-U) and lower-face action unit error (AUE-L), with AUE-L specifically assessing the similarity of individual talking motions \cite{li2024talkinggaussian}.

\begin{figure*}[b]
\centering
\includegraphics[width=\textwidth]{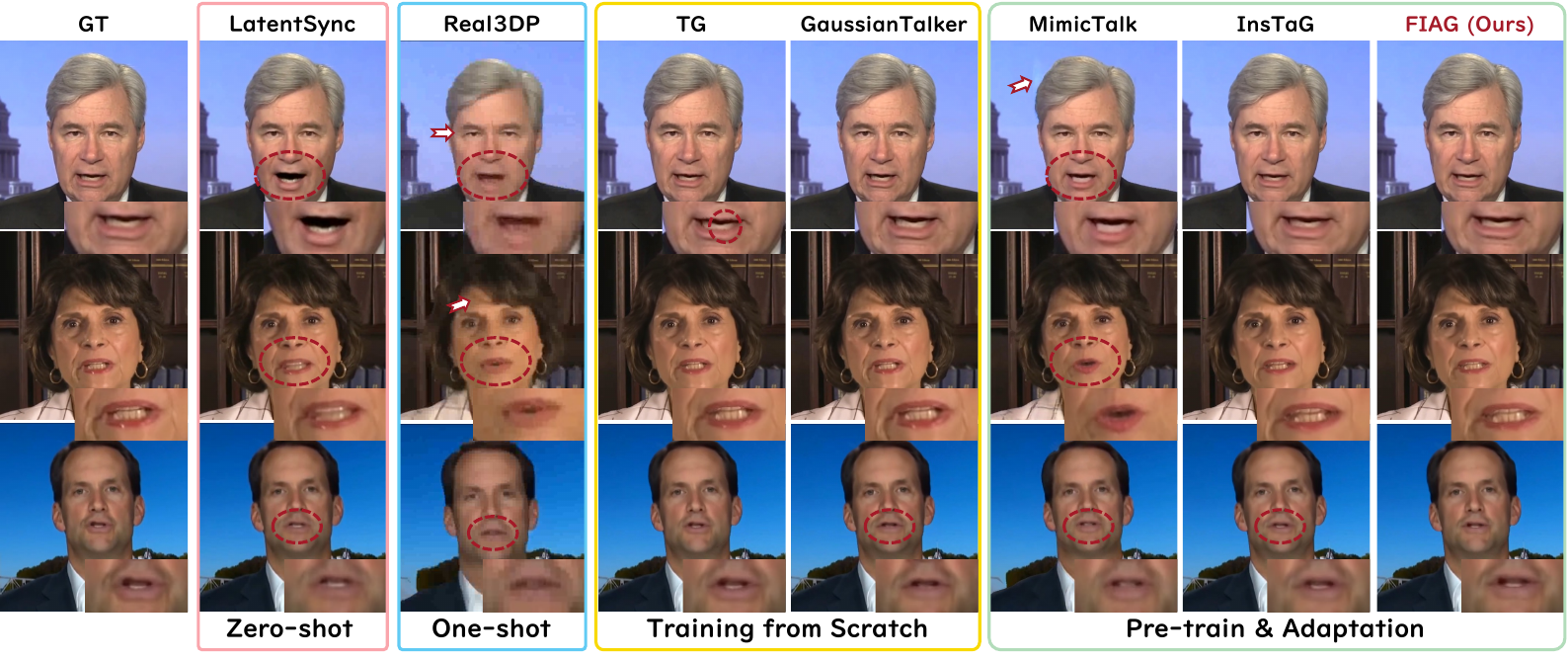}
\caption{\textbf{Visual comparison in lip image quality.} The bottom right corner presents a magnified view of the mouth region to reveal finer details. The red dotted ellipse and arrow indicate some errors.}
\label{fig6}
\end{figure*}

\subsection{Quantitative Results}

1) \textit{Self-reconstruction with a few pieces of footage}: TABLE \ref{tab1} summarizes the self-reconstruction results obtained using only five seconds of training data. Our approach consistently achieves superior performance in terms of visual fidelity, lip synchronization accuracy, and generation efficiency. Zero-shot methods, which leverage large-scale multi-speaker pretraining, show competitive lip-sync capabilities but are often limited by lower image quality and slower generation speeds. Conversely, one-shot methods demonstrate strong capabilities in personalized facial feature editing, yet their effectiveness in self-reconstruction remains limited. Methods trained from scratch naturally excel in identity reconstruction and facial motion representation, but face challenges in generalizing efficiently to novel identities. Notably, approaches such as DFRF \cite{shen2022learning}, RAD-NeRF \cite{tang2022real}, and ER-NeRF  \cite{li2023efficient} exhibit diminished reconstruction quality when training data is scarce. Overall, in comparison to other pretraining and adaptation-based frameworks, our method delivers the most balanced and robust performance across all evaluated metrics. 
2) \textit{Self-reconstruction with increasing training frames:}
While expanding the training dataset but maintaining a constant number of training or finetuning epochs, FIAG consistently achieves superior visual quality alongside relatively excellent lip synchronization performance, as shown in TABLE \ref{tab2}. 3) \textit{External audio-driven generation:} TABLE \ref{tab3} reports lip-sync performance in a cross-domain setting, where the driving audio (A) originates from a speaker different from the target identity (I). While methods like ER-NeRF \cite{li2023efficient} and GaussianTalker \cite{cho2024gaussiantalker} achieve reasonable results in self-reconstruction tasks, their performance significantly deteriorates under cross-domain generation conditions. In contrast, our approach demonstrates notable robustness when handling variations across gender and language domains, highlighting its generalization capability in diverse and challenging scenarios.

\begin{figure*}[t]
\centering
\includegraphics[width=\textwidth]{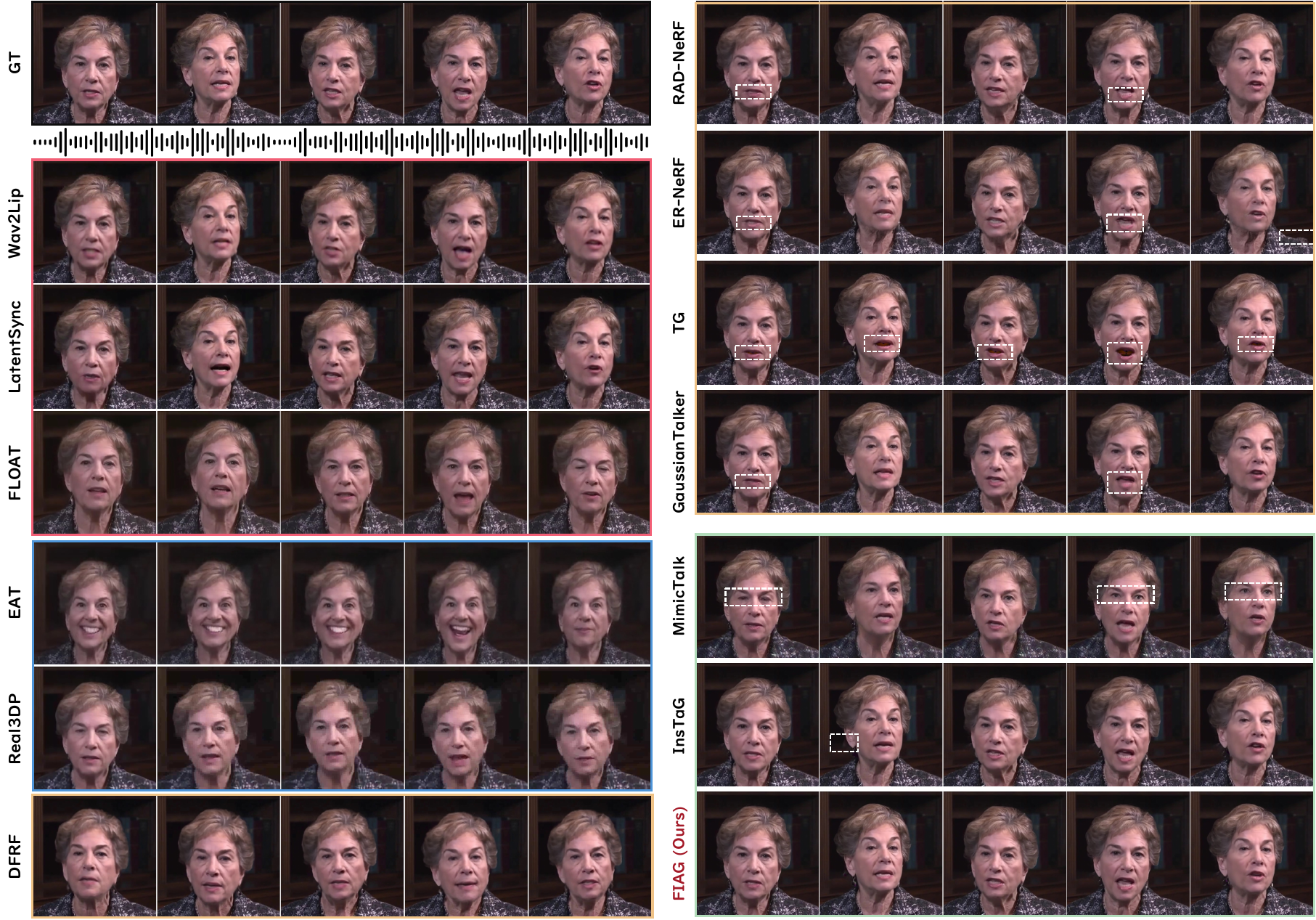}
\caption{\textbf{Visual comparison in lip synchronization accuracy.} The white dashed rectangle highlights areas containing errors.}
\label{fig5}
\end{figure*}

\begin{table*}[t]
\centering
\caption{Comparison of various methods under self-reconstruction configuration  as the amount of training data increases.}
\label{tab2}
\renewcommand{\arraystretch}{1.4}
\begin{tabular}{lcccccccccc}
\toprule
\multirow{2}{*}{\centering\vspace*{-6pt} Methods} & \multirow{2}{*}{\centering\vspace*{-6pt}\centering Config.} & \multicolumn{3}{c}{5s} & \multicolumn{3}{c}{10s} & \multicolumn{3}{c}{20s} \\
\cmidrule(lr){3-5} \cmidrule(lr){6-8} \cmidrule(lr){9-11}
 & & PSNR $\uparrow$ & LMD $\downarrow$ & Sync-C $\uparrow$ & PSNR $\uparrow$ & LMD $\downarrow$ & Sync-C $\uparrow$ & PSNR $\uparrow$ & LMD $\downarrow$ & Sync-C $\uparrow$ \\
\midrule
Ground Truth & - & - & - & 2.023 & - & - & 1.289 & - & - & 1.183\\
\hline
RAD-NeRF \cite{tang2022real} & \multirow{4}{*}{\centering \makecell{Training \\ from \\ Scratch}} & 26.4932 & 2.3206 & 1.340 & 24.3868 & 2.3727 & 0.675 & 23.5662 &  2.5090 & 0.961 \\
ER-NeRF \cite{li2023efficient} & & 25.7446 & 2.2220 & 1.764 & 25.0322 & 2.3085 & 1.111 & 24.2554 & 2.3750 & 0.584 \\
TalkingGaussian \cite{li2024talkinggaussian} & & \underline{36.2547} & 2.1397 & 1.710 & \underline{36.4346} & \textbf{2.1534} & \underline{1.208} & 34.6742 & 2.2332 & \underline{1.035} \\
GaussianTalker \cite{cho2024gaussiantalker} & & 34.7823 & 2.2982 & 1.718 & 34.7930 & \underline{2.1887} & \textbf{1.233} & 34.4238 & \textbf{2.2076} & \textbf{1.108} \\
\hline
MimicTalk \cite{ye2024mimictalk} & \multirow{3}{*}{\centering \makecell{Pre-train \& \\ Adaptation}}  & 24.1083 & 2.5822 & 1.703 & 23.9187 & 3.1694 & 1.139 & 23.2060 & 3.3743 & 0.867\\
InsTaG \cite{li2025instag} & & 35.8987 & 2.1945 & 1.741 & 32.1333 & 2.4364 & 1.137 & 33.2340 & 2.5832 & 0.592 \\
\textbf{FIAG (Ours)} & & \textbf{37.6899} & \textbf{2.0954} & \textbf{1.909} & \textbf{36.5952} & 2.2122 & 1.206 & \textbf{35.7557} & \underline{2.2184} & 0.905\\
\bottomrule
\end{tabular}
\end{table*}

\subsection{Qualitative Analysis}
Fig. \ref{fig5} and Fig. \ref{fig6} illustrates the talking head sequences and critical details generated by various framework-based methods under the self-reconstruction configuration. Methods relying on single-image generation, such as FLOAT \cite{ki2024float}, EAT \cite{gan2023efficient}, and Real3DPortrait \cite{yereal3d}, struggle to effectively accomplish self-reconstruction. Approaches that generate the mouth region separately after cropping, exemplified by Wav2Lip \cite{prajwal2020lip} and LatentSync \cite{li2024latentsync}, often result in mouth areas of inferior quality, exhibiting discontinuities with the surrounding facial regions. In contrast, reconstruction and rendering-based methods employing a similar segmentation strategy substantially enhance the continuity of facial resolution, although occasional failures in intraoral generation, exemplified by TG, may still occur. NeRF-based methods continue to demonstrate limitations in accurately capturing subtle motion nuances. While 3D GS-based methods have significantly advanced the overall fidelity of generated sequences, they remain susceptible to sporadic motion artifacts and inaccuracies in mouth articulation. Benefiting from the accumulation of generalized head topology and motion patterns acquired during pretraining, FIAG achieves superior performance in both rendering fidelity and motion precision with minimal finetuning.

\begin{table}[htbp]
\centering
\renewcommand{\arraystretch}{1.5}
\caption{Comparison of various methods in terms of lip-sync accuracy under cross-domain configuration.}
\label{tab3}
\begin{tabular}{lcccc}
\toprule
\multirow{2}{*}{\centering\vspace*{-6pt} Methods} & \multicolumn{2}{c}{\textit{I: Male, A: English}} & \multicolumn{2}{c}{\textit{I: Female, A: German}} \\
\cmidrule(r){2-3} \cmidrule(l){4-5}
 & Sync-E $\downarrow$ & Sync-C $\uparrow$ & Sync-E $\downarrow$ & Sync-C $\uparrow$ \\
\midrule
ER-NeRF \cite{li2023efficient} & 11.834 & 2.010 & 12.883 & 0.146 \\
TalkingGaussian \cite{li2024talkinggaussian} & \underline{10.291} & \underline{3.724} & 11.937 & 0.701 \\
GaussianTalker \cite{cho2024gaussiantalker} & 12.404 & 1.783 & 12.670 & 0.460 \\
InsTaG \cite{li2025instag} & 10.541 & 3.638 & \underline{11.778} & \underline{0.764} \\
\midrule
\textbf{FIAG (Ours)} & \textbf{10.133} & \textbf{4.102} & \textbf{11.776} & \textbf{0.839} \\
\bottomrule
\end{tabular}
\end{table}

\subsection{Ablation Study}
\begin{figure}[t]
\centering
\includegraphics[width=\linewidth]{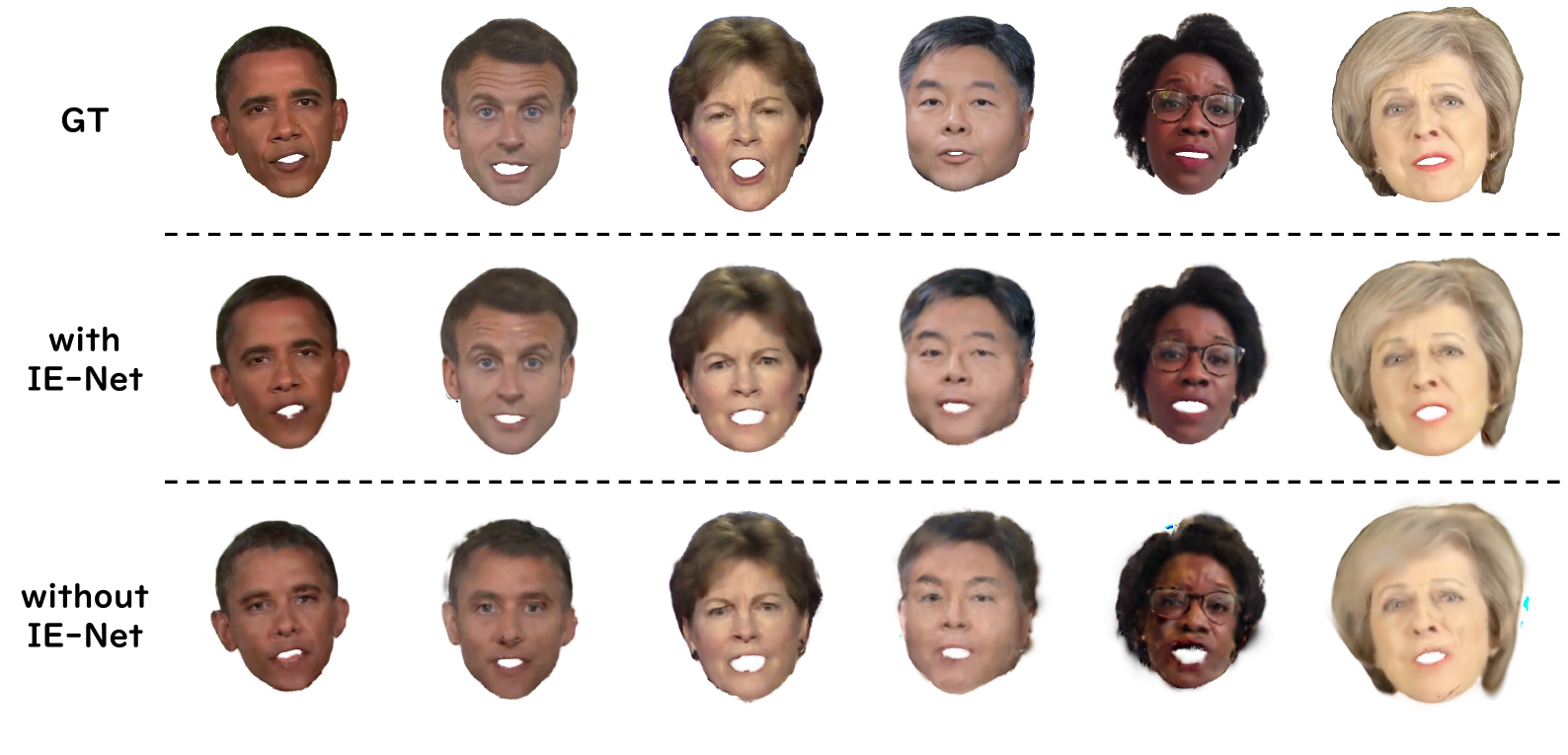}
\caption{\textbf{Visual comparison with and without identity embedding module.} The absence of identity embedding will cause serious identity conflicts.}
\label{fig7}
\end{figure}

To validate the effectiveness of our proposed method, we conducte ablation studies on the Global Gaussian Field (GGF) and comparative experiments with the Exclusive Gaussian Field (EGF). The results are presented in TABLE \ref{tab4}. The experiments demonstrate that in the absence of IM-Net, the GGF still outperforms the EGF, indicating the efficacy of our PAA architecture. However, when Identity Embedding Module (IE-Net) is removed, the performance of GGF deteriorates below that of EGF, suggesting that the identity-conflicted GGF fails to capture the common facial features and may even have adverse effects. Additionally, we visualize the rendering results of GGF with and without IE-Net as shown in Fig. \ref{fig7}.

\begin{table}[htbp]
  \centering
  \renewcommand{\arraystretch}{1.3} 
  \caption{Ablation experiment under  5s self-reconstruction configuration.}
\label{tab4}
  \begin{tabular}{cc|c|cccc}
    \toprule
    \multicolumn{2}{c|}{GGF} & \multirow{2}{*}{EGF} & \multicolumn{4}{c}{Metrics} \\
    IE-Net & IM-Net &  & PSNR $\uparrow$ & LPIPS $\downarrow$ & LMD $\downarrow$ & Sync-C $\uparrow$ \\
    \midrule
     &  & \checkmark &  36.2771 & 0.037 & 3.239 & 1.648 \\
    \checkmark &  &    & 36.6865 & 0.037 & 2.088 & 1.871\\
     & \checkmark &    &  35.2088 & 0.043 & 2.409 & 1.575\\
    \midrule
    \checkmark & \checkmark  &  & \textbf{37.6899} & \textbf{0.030} & \textbf{2.095} & \textbf{1.909} \\
    \bottomrule
  \end{tabular}
\end{table}

\section{Conclusion}
This paper proposes a method that utilizes a Global Gaussian Field to synthesize personalized 3D talking heads through finetuning with limited data. To the best of our knowledge, the method addresses the identity conflict problem using a non-physical isolation approach for the first time. Extensive experiments have been conducted to validate the effectiveness of our method. All the identities used in this article are solely for scientific research purposes.

\newpage

\title{\LARGE Few-Shot Identity Adaptation for 3D Talking Heads via \\ Global Gaussian Field}
\author{\Large Supplementary Material}
\maketitle

\section*{Overview}
In the supplementary material, we first provide extensive additional experimental results in Section VI. Section VII presents the visualization results, followed by a discussion of the limitations of our method and potential future research directions in Section VIII. Finally, Section IX contains relevant legal and ethical statements.

\section{Additional Experiments}
\subsection{Detailed Experimental Setup}

\subsubsection{Dataset}
In our experiment, the pretraining set is constructed by combining five long videos ("Obama", "\textbf{Shaheen}", "Macron", "Lieu", "\textbf{May}") from InsTaG \cite{li2025instag} with five suitably long videos ("Cory Gardner", "Gerry Connolly", "\textbf{Karen Bass}", "\textbf{Jacky Rosen}", "\textbf{Lauren Underwood}") selected from the HDTF \cite{zhang2021flow} dataset, where bolded names denote female subjects. The test set comprises five additional long videos ("\textbf{Jan Schakowsky}", "Jim Himes", "\textbf{Lucille Roybal-Allard}", "Sheldon Whitehouse") from the HDTF dataset, with no identity overlap between pretraining and test sets. All videos have an average duration of approximately five minutes, a frame rate of 25 FPS, feature centered subjects, and are uniformly scaled and cropped to a resolution of 512 $\times$ 512 pixels. Subsequent preprocessing follows established protocols as described in prior works \cite{tang2022real, li2023efficient, cho2024gaussiantalker, li2024talkinggaussian, li2025instag}. The videos utilized in this study are sourced from publicly available online content. To ensure the protection of personal privacy, the dataset exclusively comprises public persons.

\subsubsection{Evaluation Metric}
Following the previous evaluation paradigm, the synthesized talking heads are appraised in terms of visual quality and lip-sync performance. Visual quality is measured using PSNR, LPIPS \cite{zhang2018unreasonable}, and SSIM \cite{wang2004image} metrics. Lip-sync evaluation considers both the synchronization between audio and mouth movements and the accuracy of mouth motion states, quantified by Sync-C \cite{chung2017lip, chung2017out}, LMD \cite{chen2018lip}, and AUE metrics, respectively.

\subsubsection{Comparison Baselines}
To demonstrate the superiority of our method, we conduct comparisons with state-of-the-art and classical models across different frameworks. For zero-shot, we compare against GAN-based Wav2Lip \cite{prajwal2020lip}, DDPM-based LatentSync \cite{li2024latentsync}, and Flow Matching-based FLOAT \cite{ki2024float}. For one-shot, EAT \cite{gan2023efficient} and Real3DPortrait \cite{yereal3d} (Real3DP) are selected. For training from scratch, five representative models, DFRF \cite{shen2022learning}, RAD-NeRF \cite{tang2022real}, ER-NeRF \cite{li2023efficient}, TalkingGaussian \cite{li2024talkinggaussian} (TG), GaussianTalker \cite{cho2024gaussiantalker}, are evaluated, as shown in TABLE \ref{tab1}. Within the same framework, comparisons included MimicTalk \cite{ye2024mimictalk} and InsTaG \cite{li2025instag}.

\begin{figure}[t]
\centering
\includegraphics[width=\linewidth]{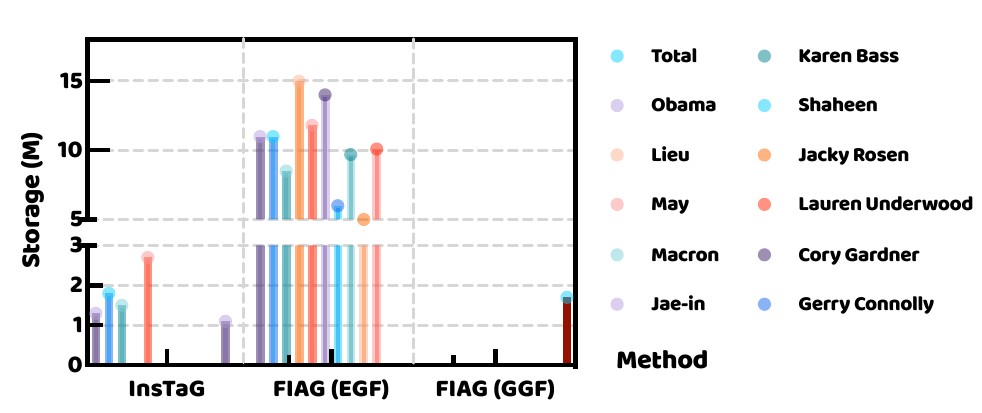}
\caption{The storage size of the Gaussian field for each identity.}
\label{fig8}
\end{figure}

\subsubsection{Implementation Details}
Unlike prior methods that train each branch separately followed by fusion training, our work adopts a joint training strategy that accomplishes the process in a single training. The pretraining procedure is divided into four stages as shown in Algorithm \ref{alg1}: warm-up, identity embedding, motion learning, and fusion. The loss function during the warm-up stage comprises the primary loss $\mathcal{L}$ of both face and mouth branches along with various regularization terms; the identity embedding and motion learning stages further incorporate the losses $\mathcal{L}_t$ and $\mathcal{L}_c$, respectively, based on the warm-up stage loss; the fusion stage loss is assigned to $\mathcal{L}$. The adaptation procedure contains no loss associated with identity discrimination as shown in Algorithm \ref{alg2}. Unless otherwise specified, all experiments utilize audio features extracted by the DeepSpeech \cite{hannun2014deep} model.

\begin{algorithm}[h]
\caption{Pretraining Process of FIAG.}
\label{alg1}
{\renewcommand{\baselinestretch}{1.3}\selectfont
\begin{algorithmic}
\STATE Definition: \small{Global Gaussian Field ($\theta_c$), Identity Embedding ($\textit{IE-Net}$), Identity Motion ($\textit{IM-Net}$), Univeral Motion ($\textit{UM-Net}$)}
\STATE \textbf{Input}: $id_{1,...,n},\ E$ epochs\STATE \textbf{Output}: $\{\theta_c,\ \textit{IE-Net},\ \textit{UM-Net}\}.pth$
\STATE Initialize $\theta_c$
\STATE \textbf{for} $e$ in $E$ epochs:
\STATE \hspace{0.5cm} random frame $I_i^{t}$ in ($id_{1,...,n}$)
\STATE \hspace{0.5cm} $\Delta \theta_i = \textit{IE-Net}(\theta_c, I_i^{t})$, $\theta_i = \Delta \theta_i + \theta_c$
\STATE \hspace{0.5cm} \textbf{if} \textit{e} in [\textit{warm-up, identity embedding, motion learning}]:
\STATE \hspace{1.0cm} $\mathcal{L}(\mathcal{R}(\theta_i^f), I_i^{t}) + (\mathcal{R}(\theta_i^m), I_i^{t}) + \textit{Norm}(\mathcal{R}(\theta_i))$ 
\STATE \hspace{0.5cm} \textbf{if} \textit{e} in \textit{identity embedding} stage:
\STATE \hspace{1.0cm} $\mathcal{L}_t(\Delta \theta_i^a, \Delta \theta_i^p, \Delta \theta_\phi^n)$ 
\STATE \hspace{0.5cm} \textbf{if} \textit{e} in \textit{motion learning} stage:
\STATE \hspace{1.0cm} $\delta_i^f = \textit{IM-Net}(\theta_i^f, C)$, $\theta_i^f = \theta_i^f + \delta_i^f$
\STATE \hspace{1.0cm} $\mathcal{L}_c(\delta_i^f, \delta_\phi^f)$ 
\STATE \hspace{1.0cm} Face: $\delta_i^f = \textit{UM-Net}(\theta_i^f, C)$, $\theta_i^f = \theta_i^f + \delta_i^f$
\STATE \hspace{1.0cm} $\mathcal{E} = \textit{Top-k Selector}(\theta_i^f\{\mu_i\})$  
\STATE \hspace{1.0cm} Mouth: $\delta_i^m = \textit{UM-Net}(\theta_i^m, C, \mathcal{E})$, $\theta_i^m = \theta_i^m + \delta_i^m$
\STATE \hspace{0.5cm} \textbf{if} \textit{e} in \textit{fusion} stage:
\STATE \hspace{1.0cm} $\mathcal{L}(\mathcal{R}(\theta_i), I_i^{t})$ 
\STATE \hspace{0.5cm} save checkpoint
\STATE \textbf{end}
\end{algorithmic}
}
\end{algorithm}

\begin{algorithm}[H]
\caption{Adaptation Process of FIAG.}
\label{alg2}
{\renewcommand{\baselinestretch}{1.3}\selectfont
\begin{algorithmic}
\STATE \small Input: $id_\varphi,\ \{\theta_c,\ \textit{IE-Net},\ \textit{UM-Net}\}.pth,\ E$ epochs
\STATE Output: $\{\theta_\varphi,\ \textit{IE-Net},\ \textit{UM-Net}\}.pth$
\STATE Initialize $\theta_c,\  \textit{IE-Net},\ \textit{UM-Net} \leftarrow checkpoint.pth, \ \textit{IM-Net} \leftarrow pop\ \textit{IM-Pool}$
\STATE \textbf{for} $e$ in $E$ epochs:
\STATE \hspace{0.5cm} random frame $I_{\varphi}^{t}$ in $id_{\varphi}$
\STATE \hspace{0.5cm} $\Delta \theta_{\varphi} = \textit{IE-Net}(\theta_c, I_{\varphi}^{t})$, $\theta_{\varphi} = \Delta \theta_{\varphi} + \theta_c$
\STATE \hspace{0.5cm} \textbf{if} \textit{e} in [\textit{warm-up, identity embedding, motion learning}]:
\STATE \hspace{1.0cm} $\mathcal{L}(\mathcal{R}(\theta_{\varphi}^f), I_{\varphi}^{t}) + (\mathcal{R}(\theta_{\varphi}^m), I_{\varphi}^{t}) + \textit{Norm}(\mathcal{R}(\theta_{\varphi}))$ 
\STATE \hspace{0.5cm} \textbf{if} \textit{e} in \textit{motion learning} stage:
\STATE \hspace{1.0cm} $\delta_{\varphi}^f = \textit{IM-Net}(\theta_{\varphi}^f, C)$, $\theta_{\varphi}^f = \theta_{\varphi}^f + \delta_{\varphi}^f$
\STATE \hspace{1.0cm} Face: $\delta_{\varphi}^f = \textit{UM-Net}(\theta_i^f, C)$, $\theta_{\varphi}^f = \theta_{\varphi}^f + \delta_{\varphi}^f$
\STATE \hspace{1.0cm} $\mathcal{E} = \textit{Top-k Selector}(\theta_{\varphi}^f\{\mu_{\varphi} \})$  
\STATE \hspace{1.0cm} Mouth: $\delta_{\varphi}^m = \textit{UM-Net}(\theta_{\varphi}^m, C, \mathcal{E})$, $\theta_{\varphi}^m = \theta_i^m + \delta_{\varphi}^m$
\STATE \hspace{0.5cm} \textbf{if} \textit{e} in \textit{fusion} stage:
\STATE \hspace{1.0cm} $\mathcal{L}(\mathcal{R}(\theta_{\varphi}), I_{\varphi}^{t})$ 
\STATE \hspace{0.5cm} save checkpoint
\STATE \textbf{end}
\end{algorithmic}
}
\end{algorithm}

\subsection{Comparative Experiments on Gaussian Utilization}
\begin{figure}[t]
\centering
\includegraphics[width=\linewidth]{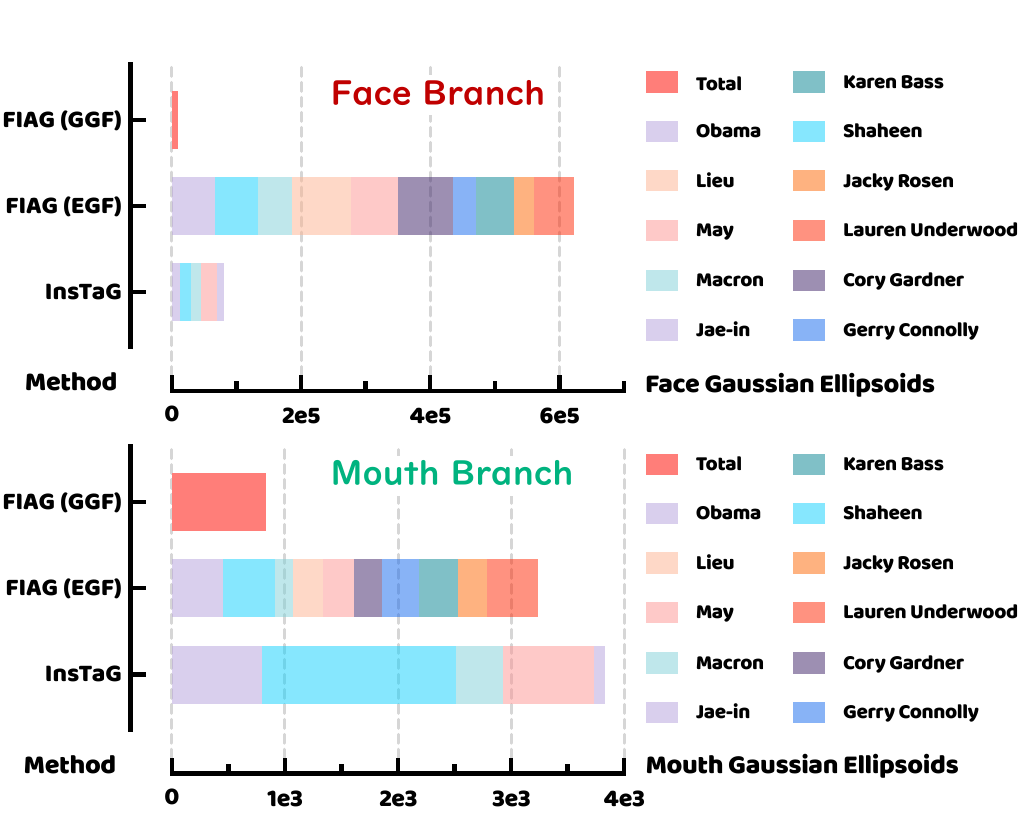}
\caption{Comparison of the number of Gaussian ellipsoids constituting the Gaussian fields on the face and mouth branches.}
\label{fig9}
\end{figure}

Fig. \ref{fig8} presents a statistical comparison of the Gaussian field file sizes generated by InsTaG \cite{li2025instag} and FIAG using the Global Gaussian Field (GGF) and Exclusive Gaussian Field (EGF) approaches, respectively, after pretraining. Fig. \ref{fig9}, on the other hand, illustrates the statistics of the number of Gaussian ellipsoids required to construct the Gaussian field for each identity. TABLE \ref{tab5} presents the memory consumption of FIAG and InsTaG at different stages and for varying numbers of identities, where N-k denotes pretraining conducted on k identities. InsTaG is pretrained on five identities: "Obama", "Shaheen", "Jae-in", "May", and "Macron". FIAG (EGF) employs an exclusive Gaussian field scheme, allocating a separate Gaussian field for each identity. FIAG (GGF), utilizing a global Gaussian field scheme, is pretrained on ten identities: "Obama", "Shaheen", "Lieu", "May", "Macron", "Karen Bass", "Jacky Rosen", "Lauren Underwood", "Cory Gardner", and "Gerry Connolly", only creating a single global Gaussian field file.

In the context of multi-identity pretraining, FIAG (GGF) utilizes approximately 10,000 Gaussian ellipsoids to complete the pretraining of ten identities, whereas InsTaG requires around 80,000 Gaussian ellipsoids to pretrain only five identities. This indicates that the utilization efficiency of InsTaG is merely one-sixteenth that of FIAG (GGF), demonstrating the remarkable ellipsoid reuse capability of our global Gaussian field. This advantage is further reflected in storage consumption, with the GGF file size being 1.7 MB compared to 8.4 MB for InsTaG. Moreover, when compared to the EGF approach, this benefit is even more pronounced: EGF requires a total of 662,000 ellipsoids for ten identities, resulting in a Gaussian ellipsoid reuse rate of up to 98.5\% for GGF. 

Although both methods utilize the exclusive Gaussian field (EGF) approach, FIAG (EGF) differs from InsTaG primarily by the incorporation of the IE-Net module. Notably, FIAG (EGF) requires a substantially larger number of Gaussian ellipsoids to represent the Gaussian field for each identity compared to InsTaG. This discrepancy can be attributed to the distribution of Gaussian ellipsoids observed in Fig. \ref{fig3}, where a pronounced clustering occurs around the head contour region. Such concentrated allocation of ellipsoids in this area accounts for the increased total number employed by FIAG (EGF). The underlying reason for this clustering stems from the fundamental design philosophy of the global Gaussian field. The global Gaussian field aims to comprehensively capture the common features shared among different identities. While the spatial arrangement of facial features tends to be relatively consistent across individuals, the head contour (including hair) exhibits significantly greater variability. Consequently, the majority of Gaussian ellipsoids are concentrated in this contour region, enabling the IE-Net to effectively model the diverse head shapes of different identities through minimal residual offsets. This strategy facilitates efficient representation of individual-specific variations while leveraging shared structural commonalities. Although EGF replaces GGF, the IE-Net module remains incorporated, thereby preserving this characteristic and ultimately leading to the observed outcome. 

\begin{table}[htbp]
  \centering
  \renewcommand{\arraystretch}{1.5}
  \caption{VRAM usage (MB) for different stages and quantities of identities}
  \label{tab5}
  \begin{tabular}{lcc|ccccc}
    \toprule
    \multirow{2}{*}{Method} & \multirow{2}{*}{Infer.} & \multirow{2}{*}{Adapt.} & \multicolumn{5}{c}{Pretraining}
    \\ & & & N-2 & N-4 & N-6 & N-8 & N-10 \\
    \midrule
    InsTaG$^*$ & 1826 & 2214 & 2188 & 2204 & - & - & - \\
    \textbf{FIAG} & 1900 & 2264 & 2302 & 2294 & 2310 & 2318 & 2330   \\
    \bottomrule
  \end{tabular}

\end{table}

Furthermore, the number of Gaussian ellipsoids is also related to memory consumption. As shown in TABLE \ref{tab5}, although the GGF in FIAG contains fewer ellipsoids than the Gaussian field required for any individual identity in InsTaG, the memory usage of FIAG is comparable to that of InsTaG, and even slightly higher. This can be attributed to the fact that each pretraining iteration focuses on a single identity, resulting in each EGF in InsTaG being loaded sequentially rather than simultaneously. Additionally, FIAG includes an extra IE-Net, which contributes to its slightly higher memory usage. Consequently, the relatively stable memory consumption observed in TABLE \ref{tab5} as the number of pretraining identities increases is due to different reasons for the two methods: for FIAG, it is because of the GGF, whereas for InsTaG, it is because the EGFs are loaded one at a time.

\subsection{Ablation Experiments}

To further demonstrate the effectiveness of our method, we conducte additional ablation experiments on several key components within the motion field as well as the loss function used for identity disentanglement. The results are presented in TABLE \ref{tab6}. When all modules responsible for generating personalized motion (IM-Net and 
$\mathcal{E}$) are removed, the model exhibits a significant decline in lip-sync performance metrics (LMD drop 8\%, Sync-C drop 7.7\%). This indicates that IM-Net plays a crucial role in personalizing motion adjustments within the facial branch. Similarly, the module 
$\mathcal{E}$ is essential for effectively linking the facial and mouth branches. A comparison of the first three experiments provides a more detailed understanding of the contributions of IM-Net and the Hook \(\mathcal{E}\). When the loss function ($\mathcal{L}_t$ and $\mathcal{L}_c$) used for identity disentanglement is completely removed, the model experiences some degradation in both visual quality and lip-sync metrics; however, the impact is relatively minor, which indirectly indicates the contribution of IE-Net to identity disentanglement (compare the third one with the last one or the second with the penultimate). A comparison of the fourth, fifth, and final experiments allows for a more detailed understanding of the individual contributions of \(\mathcal{L}_t\) and \(\mathcal{L}_c\). Based on the analysis of TABLE \ref{tab4} and \ref{tab6}, we can draw the following conclusions: 1) IE-Net plays a crucial role in achieving identity disentanglement and serves as the key mechanism by which the global Gaussian field stores shared facial features across different pre-trained identities. 2) The personalized motion learning module within the motion field contributes positively to capturing identity-specific motions that differ from the generic motions stored in the common motion field. This effect is directly reflected in the metrics evaluating lip-sync accuracy and indirectly manifested in the visual quality metrics measured against real images. 3) The loss function designed to enforce identity disentanglement further enhances the independence of identities, although its impact is relatively limited.

\begin{table}[h]
\centering
\renewcommand{\arraystretch}{1.5}
\caption{Ablation experiment under  5s self-reconstruction configuration.}
\label{tab6}
\begin{tabular}{ccccccccc}
\toprule
IM-Net & $\mathcal{E}$ & $\mathcal{L}_t$ & $\mathcal{L}_c$ & PSNR $\uparrow$ & LPIPS $\downarrow$ & LMD $\downarrow$ & Sync-C $\uparrow$ \\
\midrule
 &  &  &  & 35.63 & 0.042 & 2.261 & 1.763 \\
\checkmark &  &  &  & 36.45 & 0.035 & 2.187 & 1.828 \\
\checkmark & \checkmark &  &  & 37.05 & 0.035 & 2.155 & 1.886 \\
\checkmark & \checkmark & \checkmark &  & 37.26 & 0.034 & 2.100 & 1.886 \\
\checkmark & \checkmark &  & \checkmark & 37.45 & 0.031 & 2.151 & 1.897 \\
\checkmark &  & \checkmark & \checkmark & 36.45 & 0.035 & 2.187 & 1.844 \\
\midrule
\checkmark & \checkmark & \checkmark & \checkmark & \textbf{37.69} & \textbf{0.030} & \textbf{2.095} & \textbf{1.910} \\
\bottomrule
\end{tabular}
\end{table}

\subsection{Audio Feature Extractor}
To ensure a fair comparison with various existing methods \cite{shen2022learning, tang2022real, li2023efficient, li2024talkinggaussian,  cho2024gaussiantalker, li2025instag}, all our previous experiments exclusively employ DeepSpeech \cite{hannun2014deep} for audio feature extraction. To further validate the effectiveness of our approach, we conducte additional experiments using multiple audio feature extractors, such as Wav2Vec 2.0 \cite{baevski2020wav2vec}, HuBERT \cite{hsu2021hubert} and AVE encoder in SyncTalk \cite{peng2024synctalk}, which is pretrained on large corpus with a lip-sync expert discriminator, consistent with prior works \cite{tang2022real, li2023efficient, li2024talkinggaussian, li2025instag}. The results are presented in TABLE \ref{tab7}. 

After changing the audio feature extractor, the longitudinal comparison of various methods in terms of evaluation metrics is summarized as follows. In terms of rendered image quality, both Wav2Vec 2.0 and AVE demonstrate improvements across all three metrics, with AVE showing a more pronounced enhancement in visual quality. Specifically, AVE achieves a 0.9\% increase in PSNR, a 3.3\% improvement in LPIPS, and a 0.2\% gain in SSIM. Similarly, RAD-NeRF \cite{tang2022real}, ER-NeRF \cite{li2023efficient}, and TalkingGaussian \cite{li2024talkinggaussian} exhibit varying degrees of visual quality enhancement when using either Wav2Vec 2.0 or AVE as the audio feature extractor. However, for InsTaG \cite{li2025instag}, the rendering quality with any of the new audio feature extractors is inferior to that obtained using DeepSpeech. In regard to lip-sync accuracy, the use of AVE yields notable improvements across three metrics (AUE, Sync-E, and Sync-C), with a 4.7\% increase observed in Sync-C. Wav2Vec 2.0 also achieves moderate gains in two metrics (Sync-E and Sync-C), including a 4.0\% improvement in Sync-C. Conversely, employing HuBERT results in declines across all metrics, which we hypothesize is due to HuBERT’s excessively large audio feature encoding space causing redundancy and noise. Similarly, both ER-NeRF and TalkingGaussian exhibit certain degrees of lip-sync accuracy enhancement when using Wav2Vec 2.0 or AVE. However, these models also experience performance degradation with HuBERT, notably with ER-NeRF showing nearly a 62\% decrease in Sync-C.

In addition to the longitudinal comparisons surrounding the replacement of audio feature extractors, we are also aware that our FIAG consistently outperforms previous SOTA models in both visual quality and lip-sync accuracy across all types of audio feature extractors. Notably, even when using HuBERT, where all models experience some degree of performance degradation, our approach still achieves the best or second-best results across all metrics. This demonstrates the strong robustness of our method to different audio feature extractors, thereby further broadening its applicability across diverse scenarios.

\begin{table*}[htbp]
\centering
\setlength{\tabcolsep}{6pt}
\renewcommand{\arraystretch}{1.5}
\caption{Comparison of various methods employing different audio extractors with self-reconstruction configuration across two critical dimensions: image quality and lip-sync accuracy.}
\label{tab7}
\begin{tabular}{l c c c c c c c c c}
\toprule
\multirow{2}{*}{\centering\vspace*{-6pt} Methods} & \multirow{2}{*}{\centering\vspace*{-6pt} Year} & \multirow{2}{*}{\centering\vspace*{-6pt}Audio Extractor} & \multicolumn{3}{c}{Image Quality} & \multicolumn{4}{c}{Lip-sync Accuracy} \\
\cmidrule(lr){4-6} \cmidrule(lr){7-10}
 & & & PSNR $\uparrow$ & LPIPS $\downarrow$ & SSIM $\uparrow$ & LMD $\downarrow$ & AUE-(U/L) $\downarrow$ & Sync-E $\downarrow$ & Sync-C $\uparrow$ \\
\midrule
Ground Truth & - & - & - & - & - & - & - & 11.321 & 2.023 \\
\midrule
RAD-NeRF \cite{tang2022real} & 2022 & \multirow{5}{*}{Wav2Vec 2.0} & 26.6573 & 0.0969 & 0.8431 & 2.3336 &  0.225 / 0.385 & 12.202 & 1.255 \\
ER-NeRF \cite{li2023efficient} &  2023 &  & 25.7881 & 0.0754 & 0.8444 & 2.1442 &  0.167 / 0.213 & 11.547 & 1.873 \\
TalkingGaussian \cite{li2024talkinggaussian} & 2024 &  & \underline{37.6302} & \textbf{0.0188} & \underline{0.9690} & \textbf{2.0198} &  \textbf{0.089} / \textbf{0.143} & \underline{11.405} & \underline{1.901} \\
InsTaG \cite{li2025instag} & 2025 &  & 35.8933 & 0.0385 & 0.9602 & 2.1262 &  0.156 / 0.267 & 11.563 & 1.860 \\
\textbf{FIAG (Ours)} & 2025 &  & \textbf{37.7095} & \underline{0.0288} & \textbf{0.9716} & \underline{2.1005} &  \underline{0.095} / \underline{0.206} & \textbf{11.376} & \textbf{1.986} \\
\midrule
ER-NeRF \cite{li2023efficient} & 2023 & \multirow{4}{*}{HuBERT} & 25.3292 & 0.0871 & 0.8314 & 2.8459 &  0.207 / 0.932 & 12.705 & 0.672 \\
TalkingGaussian \cite{li2024talkinggaussian} & 2024 &  & 35.3565 & \textbf{0.0235} & \underline{0.9596} & \underline{2.1785} &  \underline{0.128} / \underline{0.247} & 11.793 & 1.566 \\
InsTaG \cite{li2025instag} & 2025  &  & \underline{35.6461} & 0.0405 & 0.9586 & 2.2111 &  0.231 / 0.275& \underline{11.530} & \textbf{2.015} \\
\textbf{FIAG (Ours)} & 2025 &  & \textbf{37.2437} & \underline{0.0312} & \textbf{0.9696} & \textbf{2.1312} &  \textbf{0.087} / \textbf{0.238} & \textbf{11.447} & \underline{1.808} \\
\midrule
TalkingGaussian \cite{li2024talkinggaussian} & 2024 & \multirow{3}{*}{AVE} & \underline{37.8749} & \textbf{0.0183} & \underline{0.9705} & \textbf{2.0082} &  \textbf{0.077} / \textbf{0.133} & \underline{11.438} & \underline{1.965} \\
InsTaG \cite{li2025instag} & 2025 &  & 35.7481 & 0.0379 & 0.9613 & \underline{2.1457} &  0.145 / 0.283& 11.621 & 1.537 \\
\textbf{FIAG (Ours)} & 2025 &  & \textbf{38.0590} & \underline{0.0286} & \textbf{0.9731} & 2.1534 &  \underline{0.102} / \underline{0.166} & \textbf{11.350} & \textbf{2.000} \\
\bottomrule
\end{tabular}
\end{table*}

\subsection{More Training Data}
In our previous experiments, we have consistently focused on the capability of various methods in few-shot reconstruction. Although experiments with increased training data are conducted as shown in TABLE \ref{tab2}, the amount of training data remained limited. In order to further illustrate our reconstruction capabilities under the full dataset, we further conducte experiments with unrestricted access to the full training dataset as shown in TABLE \ref{tab8}. For the PAA-based methods, the training data corresponds specifically to the data used during the finetuning stage. FIAG continues to demonstrate superior performance in both image rendering quality and lip-sync accuracy compared to previous methods that either train from scratch (ER-NeRF \cite{li2023efficient} and TalkingGaussian \cite{li2024talkinggaussian}) or rely on PAA-based (InsTaG \cite{li2025instag}) approaches. With an increased amount of training data, extending the number of training iterations (from 30K to 40K) facilitates the generation of higher-quality rendered images. However, this improvement is accompanied by a certain deterioration in lip-sync accuracy.

\begin{table}[htbp]
\centering
\renewcommand{\arraystretch}{1.5}
\caption{Comparison of Various Methods Using Full Training Data.}
\label{tab8}
\begin{tabular}{l c c c c c}
\toprule
Method & Time & PSNR $\uparrow$ & SSIM $\uparrow$ & LMD $\downarrow$ & Sync-C $\uparrow$ \\
\midrule
ER-NeRF \cite{li2023efficient} & 2 hs & 23.45 & 0.762 & 3.134 & 0.587 \\
TG \cite{li2024talkinggaussian} & 36 mins & 29.63 & 0.896 & \textbf{2.968} & 0.651 \\
InsTaG \cite{li2025instag} & 19 mins & 29.82 & 0.901 & \underline{2.978} & 0.622 \\
FIAG-30K & 16 mins & \underline{30.07} & \underline{0.905} & 3.001 & \textbf{0.686} \\
FIAG-40K & 24 mins & \textbf{30.12} & \textbf{0.906} & 3.032 & \underline{0.674} \\

\bottomrule
\end{tabular}
\end{table}

\begin{figure}[h]
\centering
\includegraphics[width=\linewidth]{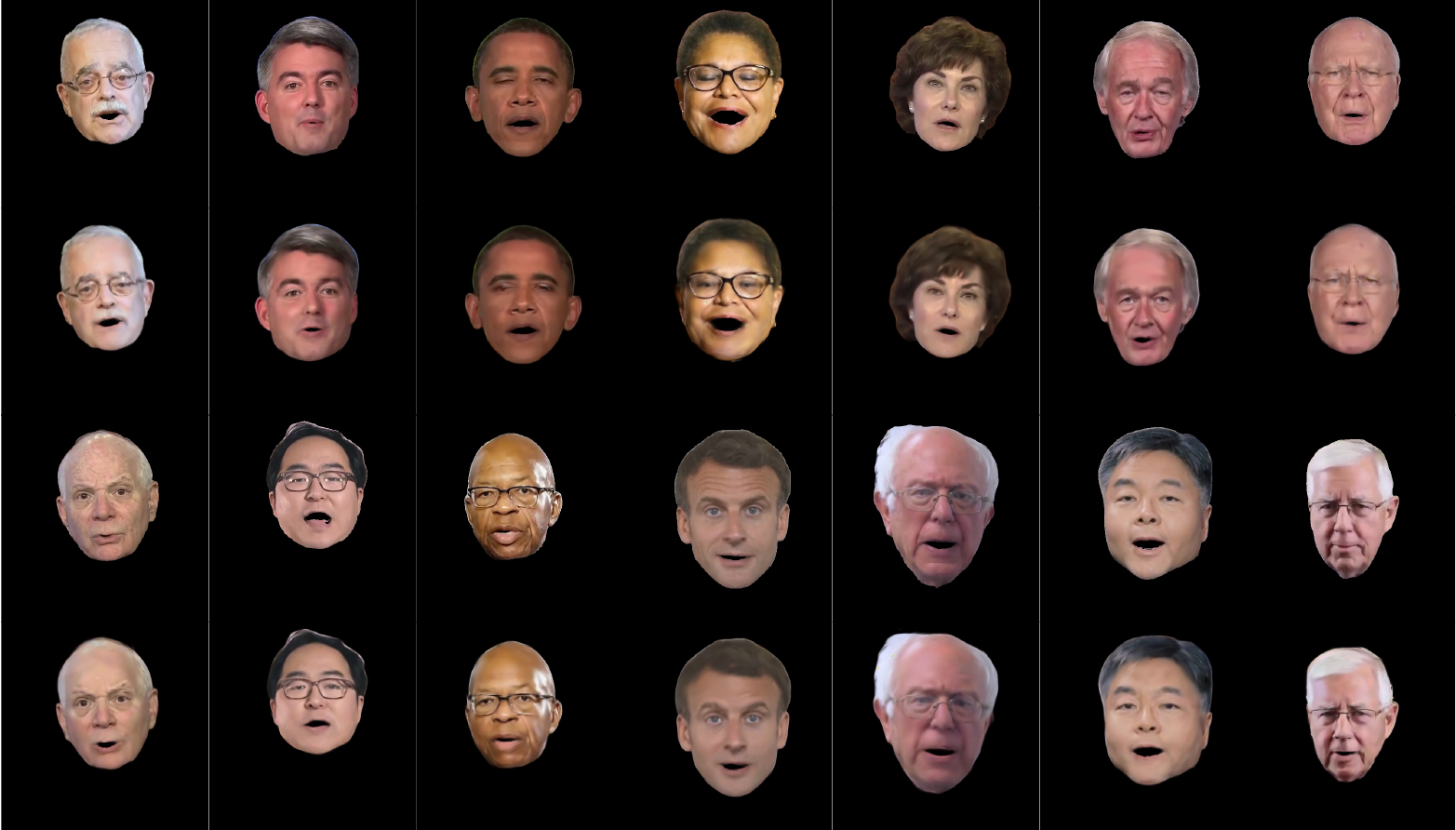}
\caption{When pretraining with 20 identities, each talking video having only a duration of 5 seconds, the GGF generates the corresponding rendering results for each identity.}
\label{fig10}
\end{figure}

\begin{figure*}[htbp]
\centering
\includegraphics[width=0.94 \textwidth]{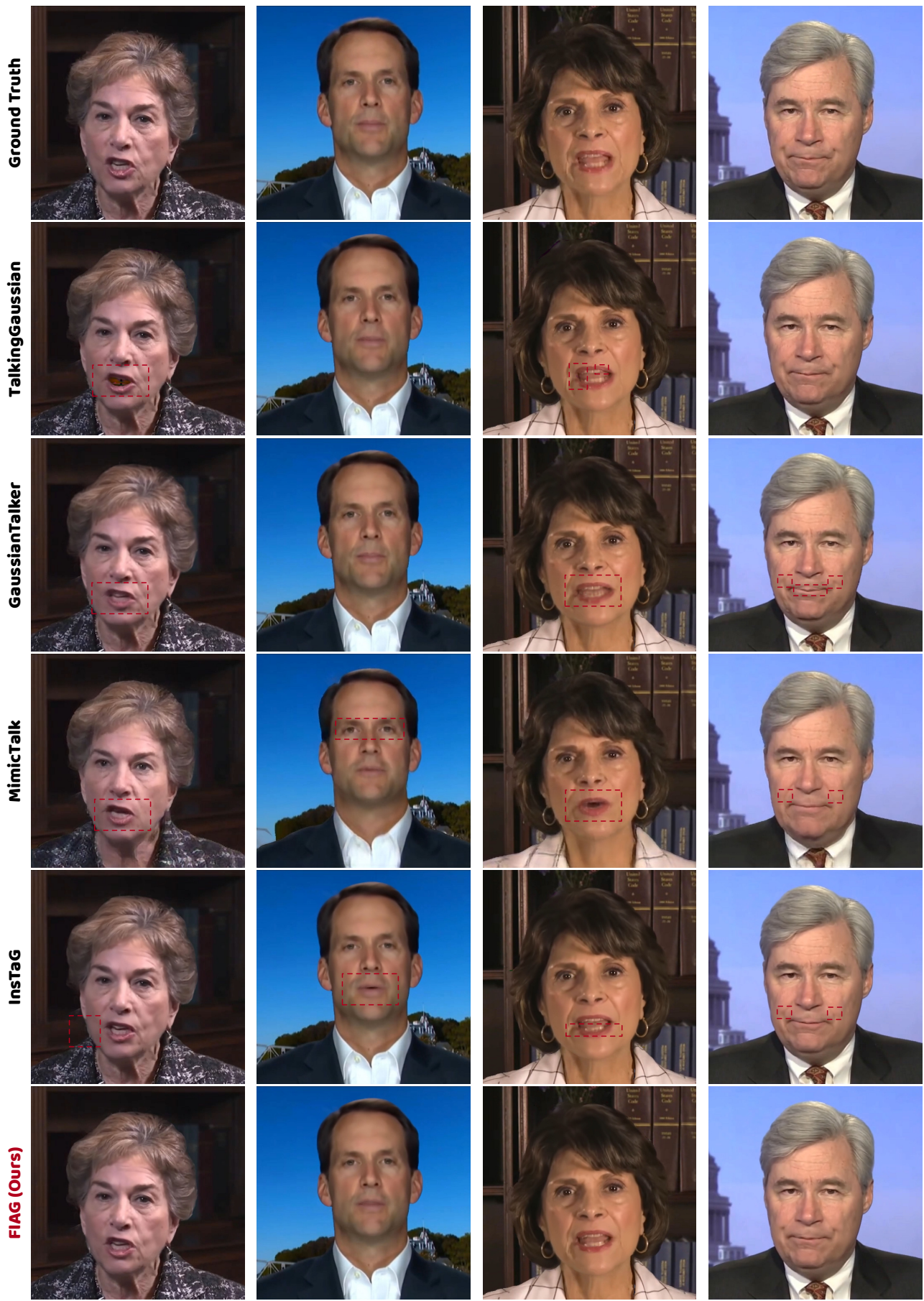}
\caption{\textbf{High-definition comparison in the quality of rendered images and lip-sync accuracy.} The red dashed rectangle highlights areas containing errors.}
\label{fig11}
\end{figure*}

\subsection{More Pretraining Identity}
With the increase in the number of training identities, identity conflicts inevitably become more severe. Moreover, the diversity of identities, such as ethnicity, gender, age, and facial features, further exacerbates these conflicts. To more comprehensively demonstrate the effectiveness of our Global Gaussian Field in addressing identity conflict issues, we doubled the amount of pretraining data while reducing the video duration for each identity from the original 5 minutes to 5 seconds. The rendering results of the Global Gaussian Field for each identity are shown in Fig. \ref{fig10} (14 identities are selected for display), where the first row corresponds to the ground truths and the second row to the rendered results.

We observe that even under the conditions of increased identities and reduced training data, the Global Gaussian Field still successfully achieves a coarse reconstruction of each identity. Although it cannot perform fine-grained reconstruction of pretrained identities to the same extent as EGF (for example, capturing more detailed features such as facial textures in “Macron” and “Ben Cardin”), the objective of the pretraining stage is primarily to achieve a rough reconstruction of identities and to learn generalized facial features.

\section{Additional Visualization}
In addition to the visualizations presented earlier, we provide higher-resolution images in the supplementary materials to more comprehensively illustrate the capacity of FIAG to generate high-quality rendered images and its superior performance in lip synchronization, as shown in Fig.\ref{fig11}. The reconstructions of the first and third identities demonstrate that our method excels at capturing intricate intraoral details. The reconstruction of the second identity highlights the capability of our approach to generate accurate lip movements even in the absence of explicit audio content. Besides, the reconstruction of the fourth identity showcases the superior performance of our method in restoring fine facial details. 

\section{Limitation and Future Work}
Our approach is built upon a framework integrating global Gaussian fields and universal motion fields, where both identity and motion representations are modeled using the PAA paradigm. Specifically, the global Gaussian field is pretrained to capture the universal facial topology, while the universal motion field learns common motion patterns shared across different identities. Through identity-specific finetuning, the model transitions from a general identity representation to a specific one by combining the universal motion field with identity-specific motion representations. Extensive experimental results have demonstrated the effectiveness and robustness of our method. Nevertheless, we acknowledge that certain limitations remain and warrant further investigation. 

The introduction of the global Gaussian field leads to a bifurcation of the static identity representation into two components: a universal identity shared through the Gaussian field, and an identity embedding module responsible for activation and prediction. In other words, unlike previous approaches \cite{li2024talkinggaussian, cho2024gaussiantalker, li2025instag} based solely on fully explicit 3D GS, our identity representation is hybrid, partially relying on the explicit 3D GS representation and partially on the implicit representation learned by the identity embedding module. This hybrid representation, compared to the fully explicit representation, exhibits a greater dependency on the training data when rendering under extreme viewing angles. In addition, although our experiments demonstrate the feasibility of pretraining FIAG across a larger number of identities, a natural question arises as to whether zero-shot capability, similar to that of generative model-based methods, can be achieved. Unfortunately, under the current framework, achieving zero-shot performance for specific identities remains unattainable.

While achieving zero-shot capability remains challenging, we have taken note of the progress made by various one-shot methods \cite{gan2023efficient, yereal3d}. These advancements motivate us to pursue the development of a comprehensive one-shot 3D talking heads synthesis framework based on the PAA paradigm in future work.

\section{Declaration}
Talking heads synthesis technology plays a significant role in various fields such as voice interaction, virtual assistants, and audiovisual production, greatly enhancing user experience and operational efficiency. However, it also poses potential risks including privacy breaches, identity infringement, and fraudulent misuse. In light of these concerns, our research aims to advance the development of talking heads synthesis technology from a scientific perspective. By releasing this work as open-source, we seek to promote transparency and collaboration within the community, while emphasizing the importance of responsible and ethical use to maximize its benefits and mitigate associated risks.

\bibliographystyle{IEEEtran}
\balance
\bibliography{root}

\end{document}